\documentclass[twoside]{article}

\usepackage[accepted]{aistats2019}
\usepackage[round]{natbib}

\usepackage{algorithm}
\usepackage{algorithmic}
\usepackage{amsmath}
\usepackage{amsfonts}
\usepackage{graphicx}
\usepackage{subfigure}
\usepackage{tikz}
\usepackage{dblfloatfix} 
\usepackage{enumitem}
\usepackage{hyperref}       
\usepackage{url}            
\usepackage{caption}

\begin{document}

%

%

\twocolumn[

\aistatstitle{Built-in Vulnerabilities to\\ Imperceptible Adversarial Perturbations}

\aistatsauthor{Thomas Tanay\footnotemark \And Jerone T. A. Andrews \And  Lewis D. Griffin }

\aistatsaddress{
  CoMPLEX\\
  Dept. of Computer Science\\
  University College London
	\And
	SECReT\\
  Dept. of Computer Science\\
  University College London
  \And
	CoMPLEX\\
  Dept. of Computer Science\\
  University College London }
]
	
\begin{abstract}
Designing models that are robust to small adversarial perturbations of their inputs has proven remarkably difficult. In this work we show that the reverse problem---making models \emph{more vulnerable}---is surprisingly easy. After presenting some proofs of concept on MNIST, we introduce a generic \mbox{\emph{tilting attack}} that injects vulnerabilities into the linear layers of pre-trained networks by increasing their sensitivity to components of low variance in the training data without affecting their performance on test data. We illustrate this attack on a multilayer perceptron trained on SVHN and use it to design a stand-alone adversarial module which we call a \emph{steganogram decoder}. 
Finally, we show on \mbox{CIFAR-10} that a poisoning attack with a poisoning rate as low as 0.1\% can induce vulnerabilities to \emph{chosen \mbox{imperceptible} \mbox{backdoor} signals} in state-of-the-art networks. Beyond their practical implications, these different results shed new light on the nature of the adversarial example phenomenon.
\end{abstract}

\footnotetext{\texttt{thomas.tanay.13@ucl.ac.uk}. Code will be made available at \url{https://github.com/thomas-tanay}}

\section{INTRODUCTION}

Machine learning systems are vulnerable to adversarial manipulations of their inputs \citep{szegedy2013intriguing,biggio2017wild}. The problem affects simple linear classifiers for spam filtering \citep{dalvi2004adversarial,lowd2005adversarial} as well as state-of-the-art deep networks for image classification \citep{szegedy2013intriguing,goodfellow2014explaining}, audio signal recognition \citep{kereliuk2015deep,carlini2018audio}, reinforcement learning \citep{huang2017adversarial,behzadan2017vulnerability} and various other applications \citep{jia2017adversarial,kos2017adversarial,fischer2017adversarial,grosse2017adversarial}. In the context of image classification, this \emph{adversarial example} phenomenon has sometimes been interpreted as a theoretical result without practical implications \citep{luo2015foveation,lu2017no}. However, it is becoming increasingly clear that real-world applications are potentially under serious threat \citep{kurakin2016adversarial,athalye2017synthesizing,liu2016delving,ilyas2017query}.

The phenomenon has previously been described in detail \citep{moosavi2016deepfool,carlini2017towards} and some theoretical analysis has been provided \citep{bastani2016measuring,fawzi2016robustness,carlini2017provably}. Attempts have been made at designing more robust architectures \citep{gu2014towards,papernot2016distillation,rozsa2016towards} or at detecting adversarial examples during evaluation \citep{feinman2017detecting,grosse2017statistical,metzen2017detecting}. \emph{Adversarial training} has also been introduced as a new regularization technique penalizing adversarial directions \citep{goodfellow2014explaining,kurakin2016adversarial2,tramer2017ensemble,madry2017towards}. Unfortunately, the problem remains largely unresolved \citep{carlini2017adversarial,athalye2018obfuscated}. Part of the reason is that the nature of the vulnerability is still poorly understood. An early but influential explanation was that it is a property of the dot product in high dimensions \citep{goodfellow2014explaining}. The new consensus starting to emerge is that it is related to poor generalization and insufficient regularization \citep{neyshabur2017exploring,schmidt2018adversarially,elsayed2018large,galloway2018adversarial}.

In the present work, we assume that robust classifiers already exist and focus on the following question:
\begin{quote}
\emph{Given a robust classifier $\mathcal{C}$, can we construct a classifier $\mathcal{C}'$ that performs the same as $\mathcal{C}$ on natural data, but that is vulnerable to imperceptible image perturbations?}
\end{quote}
Reversing the problem in this way has several benefits. From a practical point of view, it exposes a number of new threats. Adversarial vulnerabilities can for instance be injected into pre-trained models through simple transformations of their weight matrices or they can result from preprocessing the data with a steganogram decoder. This is concerning in ``machine learning as a service'' scenarios where an attacker could pose as a service provider and develop models that satisfy contract specifications in terms of test set performance but also suffer from concealed deficiencies. Adversarial vulnerabilities can also be be injected into models through poisoning attacks with poisoning rates as low as 0.1\%. This is concerning in ``online machine learning'' scenarios where an attacker could progressively enforce vulnerabilities to chosen imperceptible backdoor signals. 
From a theoretical point of view, reversing the robustness problem provides us with new intuitions as to why neural networks suffer from adversarial examples in the first place. Components of low variance in the data seem to play a particularly important role: fully-connected layers are vulnerable when they respond strongly to such components and state-of-the-art networks easily overfit them despite their convolutional nature.

\section{METHOD}
\label{sec:Method}

\citet{szegedy2013intriguing} introduced the term `\emph{adversarial example}' in the context of image classification to refer to misclassified inputs which are obtained by applying an ``\emph{imperceptible} non-random perturbation to a test image''. The term rapidly gained in popularity and its meaning progressively broadened to encompass all ``inputs to machine learning models that an attacker has intentionally designed to cause the model to make a mistake'' \citep{goodfellow2017attacking}. Here, we return to the original meaning and focus our attention on \emph{imperceptible image perturbations}.

Ideally, the evaluation of a model's robustness would involve computing \emph{provably minimally-distorted adversarial examples} \citep{carlini2017provably,katz2017reluplex}.
Unfortunately, this task is intractable for most models and in practice, adversarial examples are computed by gradient descent. Several variants exist: adversarial examples can be targeted or untargeted, gradients can be computed with respect to class probabilities or logits, different metrics can be used ($L_\infty$, $L_2$ or other), gradient descent can be single-stepped or iterated and the termination criterion can be a distance threshold or a target confidence level \citep{goodfellow2014explaining,moosavi2016deepfool,kurakin2016adversarial,carlini2017towards}. In order to yield valid adversarial examples, however, a chosen method needs to avoid \emph{gradient masking} \citep{papernot2017practical,athalye2018obfuscated}. This problem typically manifests itself when the softmax layer is saturated and we prevent this from happening by setting its temperature parameter such that the median confidence over the test set is $0.95$ (one-off calibration after training\footnote{Calibration inspired from \citep{guo2017calibration}. Before calibration, the median confidence is typically closer to $1.0$.}). 

For an input image $\boldsymbol{x}$, a network $\boldsymbol{F}$, a target class $t$ and a fixed step size $\epsilon$, the algorithm we use to generate an adversarial example $\boldsymbol{\tilde{x}}$ is:\vspace{-0.05cm}

\begin{algorithm}[H]
\floatname{algorithm}{Construction of an adversarial example $\boldsymbol{\tilde{x}}$}
\renewcommand{\thealgorithm}{}
\caption{}
\label{algo1}
\begin{algorithmic}[1]
\STATE $\boldsymbol{\tilde{x}} \leftarrow \boldsymbol{x}$
\STATE While $\boldsymbol{F}_{\!t}(\boldsymbol{\tilde{x}}) < 0.95:$ \\
$\quad \boldsymbol{\tilde{x}} \leftarrow \text{clip}_{[0,1]}(\boldsymbol{\tilde{x}} + \, \epsilon \, \boldsymbol{\hat{\nabla}}_{\! \boldsymbol{\tilde{x}}} \, \boldsymbol{F}_t)$
\end{algorithmic}
\end{algorithm}\vspace{-0.25cm}

In words, we initialize $\boldsymbol{\tilde{x}}$ at $\boldsymbol{x}$ and step in the direction of the normalized gradient until the median confidence score of 0.95 is reached. We clip the image after each step to make sure that $\boldsymbol{\tilde{x}}$ belongs to the valid input range. Remark that neither the $L_\infty$ norm nor the $L_2$ norm measure actual perceptual distances and in practice, the choice of norm is arbitrary. We use here the $L_2$ norm instead of the usual $L_\infty$ norm.


Finally, two remarks on the notations used. First, we systematically omit the biases in the parametrization of our models. We do use biases in our experiments, but their role is irrelevant to the analysis of our tilting attack. Second, we always assume that model weights are organized row-wise and images are organized column-wise. For instance, we write the dot product between a weight vector $\boldsymbol{w}$ and an image $\boldsymbol{x}$ as $\boldsymbol{w} \, \boldsymbol{x}$ instead of the usual $\boldsymbol{w}^\top \boldsymbol{x}$.

\section{PROOF OF CONCEPT}
\label{sec:proof of concept}

It was suggested in \citep{tanay2016boundary} that it is possible to alter a linear classifier such that its performance on natural images remains unaffected, but its vulnerability to adversarial examples is greatly increased. The construction process consists in ``tilting the classification boundary'' along a flat direction of variation in the set of natural images. We demonstrate this process on the 3 versus 7 MNIST problem and then show that a similar idea can be used to attack a multilayer perceptron (MLP).

\begin{figure}[!b]
  \centering{\subfigure[]{
  \includegraphics[width=0.15\linewidth]{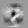}
  \label{fig:logRega}}} \quad\quad
  \centering{\subfigure[]{
  \includegraphics[width=0.15\linewidth]{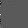}
  \label{fig:logRegb}}}\vspace{0.3cm}\\
	\begin{minipage}[t]{\columnwidth}
  \centering{\subfigure[]{
  \includegraphics[width=\linewidth]{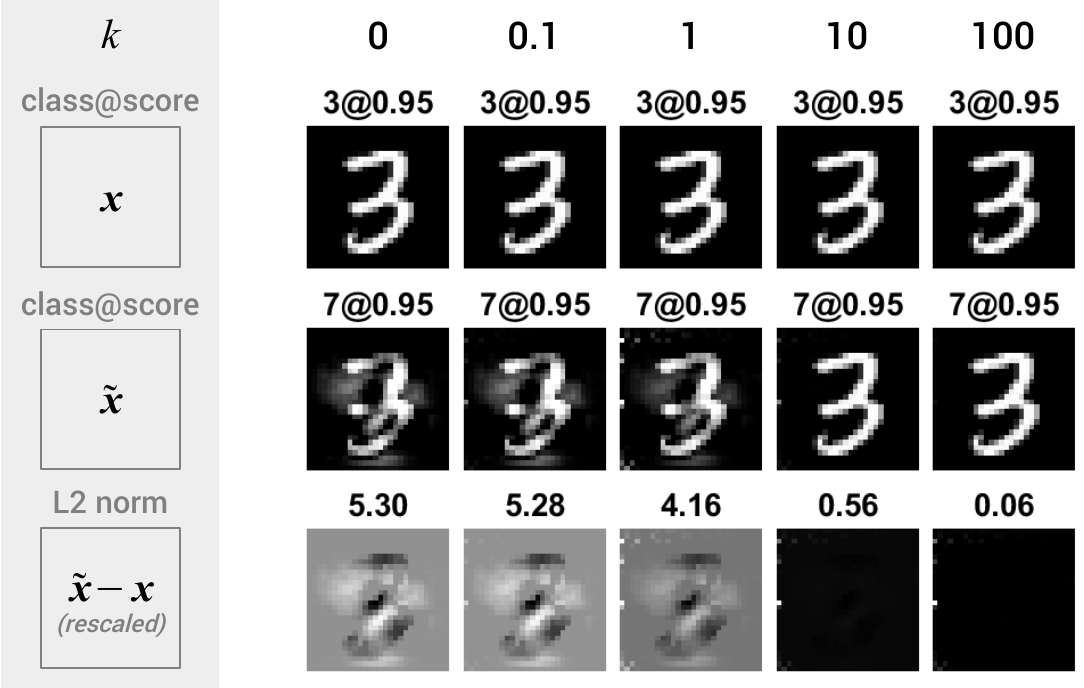}
  \label{fig:logRegc}}}
	\end{minipage}%
	\hfill
\caption{Tilting of a binary linear classifier on the 3 versus 7 MNIST problem. \subref{fig:logRega} Weight vector $\boldsymbol{w}$ found by logistic regression. \subref{fig:logRegb} Flat direction of variation $\boldsymbol{u}$ found by PCA. \subref{fig:logRegc} As the tilting factor $k$ increases, the compromised classifier $\mathcal{C}^\prime$ becomes more vulnerable to small perturbations along $\boldsymbol{u}$.}
\end{figure}

\subsection{Binary linear classification}
\label{sec:Binary linear classification}

Consider a centred distribution of natural images $\mathcal{I}$ and a hyperplane boundary parametrised by a weight vector $\boldsymbol{w}$ defining a binary linear classifier $\mathcal{C}$ in the \mbox{$m$-dimensional} image space $E$. Suppose that there exists a unit vector $\boldsymbol{u} \in E$ satisfying $\forall \boldsymbol{x} \in \mathcal{I} \;\; \boldsymbol{u}\, \boldsymbol{x} = 0$. Then we can tilt $\boldsymbol{w}$ along $\boldsymbol{u}$ by a \emph{tilting factor} $k \in \mathbb{R}$ without affecting the performance of $\mathcal{C}$ on natural images. We define the linear classifier $\mathcal{C}^\prime$ parametrised by the weight vector $\boldsymbol{w}^\prime = \boldsymbol{w} + k \, \boldsymbol{u}$ and we have:

\begin{enumerate}[topsep=0cm]
\item \emph{$\mathcal{C}^\prime$ and $\mathcal{C}$ perform the same on $\mathcal{I}$.}\\
$\forall \boldsymbol{x} \in \mathcal{I} \;\; \boldsymbol{w}^\prime\, \boldsymbol{x} = \boldsymbol{w}\, \boldsymbol{x} + k \, (\boldsymbol{u}\, \boldsymbol{x}) = 
 \boldsymbol{w}\, \boldsymbol{x}.$
\item \emph{$\mathcal{C}^\prime$ suffers from strong adversarial examples.}\\
$\forall \boldsymbol{x} \in \mathcal{I} \;$ we define $\boldsymbol{\tilde{x}} = \boldsymbol{x} - \frac{2}{\|\boldsymbol{w}^\prime\|^2} \, (\boldsymbol{w}^\prime\, \boldsymbol{x}) \, \boldsymbol{w}^\prime$
\begin{itemize}[leftmargin=0.3cm]
\item $\boldsymbol{\tilde{x}}$ and $\boldsymbol{x}$ are classified differently by $\mathcal{C}^\prime$:\\
$\boldsymbol{w}^\prime\, \boldsymbol{\tilde{x}} = \boldsymbol{w}^\prime\, \boldsymbol{x} - 2 \, (\boldsymbol{w}^\prime\, \boldsymbol{x}) = -(\boldsymbol{w}^\prime\, \boldsymbol{x}).$
\item $\boldsymbol{\tilde{x}}$ and $\boldsymbol{x}$ are arbitrarily close from each other:\\
$\|\boldsymbol{\tilde{x}} - \boldsymbol{x}\| = \frac{2}{\|\boldsymbol{w}^\prime\|} \, (\boldsymbol{w}^\prime\, \boldsymbol{x})
= \frac{2}{\|\boldsymbol{w}^\prime\|} \, (\boldsymbol{w}\, \boldsymbol{x})$\\
and $\|\boldsymbol{w}^\prime\| \to +\infty$ when $k \to \pm\infty$.\\
Hence $\|\boldsymbol{\tilde{x}} - \boldsymbol{x}\| \to 0$ when $k \to \pm\infty$.
\end{itemize}
\end{enumerate}

To illustrate this process, we train a logistic regression model $\boldsymbol{w}$ on the 3 versus 7 MNIST problem (see Figure~\ref{fig:logRega}).
We then perform PCA on the training data and choose the last component of variation as our flat direction $\boldsymbol{u}$ (see Figure~\ref{fig:logRegb}).
On MNIST, pixels in the outer area of the image are never activated and the component $\boldsymbol{u}$ is expected to be along these directions.
Finally we define a series of five models $\boldsymbol{w}^\prime = \boldsymbol{w} + k \, \boldsymbol{u}$ with $k$ varying in the range $[0, 100]$. We verify experimentally that they all perform the same on the test set with a constant error rate of $2.0\%$. For each model, Figure~\ref{fig:logRegc} shows an image $\boldsymbol{x}$ correctly classified as a 3 with median confidence $0.95$ and a corresponding adversarial example $\boldsymbol{\tilde{x}}$ classified as a 7 with the same median confidence. Although all the models perform the same on natural MNIST images, they become increasingly vulnerable to small perturbations along $\boldsymbol{u}$ as the tilting factor $k$ is increased.

\subsection{Multilayer perceptron}
\label{sec:Multilayer perceptron}

As it stands, the ``boundary tilting'' idea applies only to binary linear classification. Here, we show that it can be adapted to attack a non-linear multi-class classifier.
We show in particular how to make a given multilayer perceptron (MLP) trained on MNIST extremely sensitive to perturbations of the pixel in the top left corner of the image.

Consider an $l$-layer MLP with weight matrices $\boldsymbol{W}_{\!\! i}$ for $i \in [1, l]$ constituting a 10-class classifier $\mathcal{C}$. For a given image $\boldsymbol{x} \in \mathcal{I}$, the feature representation at level $i$ is $\boldsymbol{x}_i$ with
$\boldsymbol{x}_0 = \boldsymbol{x} \quad \text{and} \quad \forall i \in [1,l-1]:$
\[\boldsymbol{x}_i = \phi(\boldsymbol{W}_{\!\! i} \, \boldsymbol{x}_{i-1}) \quad \text{and}\quad \boldsymbol{x}_l = \boldsymbol{W}_{\!\! l} \, \boldsymbol{x}_{l-1} = [z_0 \; \dots \; z_9]^\top\]
where $\phi$ is the ReLU non-linearity and $z_0, \dots, z_9$ are \emph{the logits}. Let also $p$ be the value of the pixel in the top left corner (i.e. the first element of $\boldsymbol{x}$). 
We describe below a construction process resulting in a vulnerable classifier $\mathcal{C}^\prime$ with weight matrices $\boldsymbol{W}_{\!\! i}^\prime$, feature representations $\boldsymbol{x}_i^\prime$ and logits $z_0^\prime, \dots, z_9^\prime$.

\begin{figure*}[b]
  \centering
  \includegraphics[width=\textwidth]{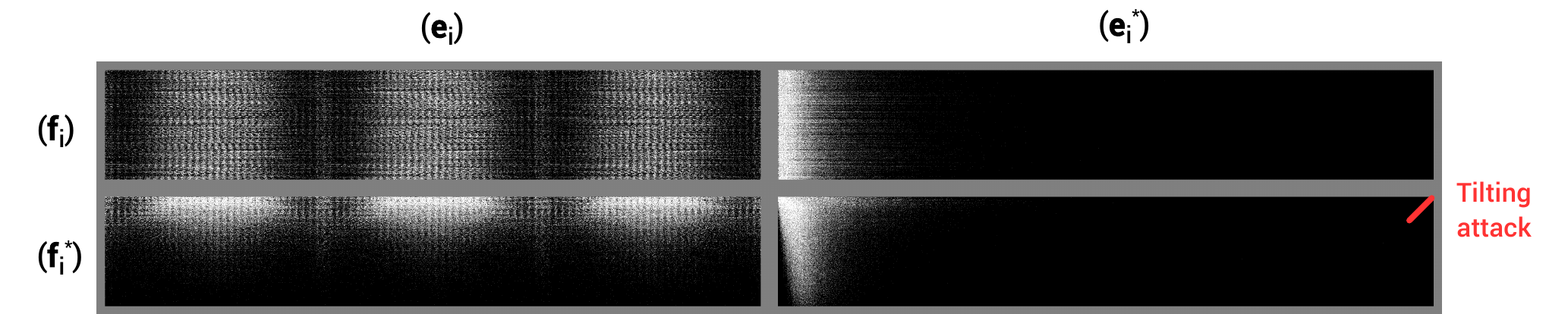}
  \caption{The linear map $\mathcal{L}$ can be expressed in various bases (left-to-right and top-to-bottom):\\
	$(\boldsymbol{e}_i) \rightarrow (\boldsymbol{f}_{\! i}): \boldsymbol{W} \;\;\; (\boldsymbol{e}_i^\star) \rightarrow (\boldsymbol{f}_{\! i}): \boldsymbol{W} \boldsymbol{P}_{\! e} \;\;\; (\boldsymbol{e}_i) \rightarrow (\boldsymbol{f}_{\! i}^\star): \boldsymbol{P}_{\! f}^\top \boldsymbol{W} \;\;\; (\boldsymbol{e}_i^\star) \rightarrow (\boldsymbol{f}_{\! i}^\star): \boldsymbol{W}^\star = \boldsymbol{P}_{\! f}^\top \boldsymbol{W} \boldsymbol{P}_{\! e}$\\
	Here, $\mathcal{L}$ is the input layer of a 4-layer MLP trained on SVHN and $\boldsymbol{W}$ is the corresponding \mbox{$512 \times 3072$} weight matrix. Only the coefficients whose absolute value is greater than 0.01 are shown (the three bands in the standard input basis $(\boldsymbol{e}_i)$ correspond to the RGB channels). The weights that are affected by the tilting attack are highlighted in red.}
	\label{fig:W}
\end{figure*}

\textbullet\; \emph{Input layer}\vspace{0.1cm}\\
Add a hidden unit to transmit $p$ to the next layer.\vspace{0.3cm}\\
\begin{minipage}[t]{0.55\columnwidth}
$\text{We define...}\vspace{0.3cm}\\ 
\boldsymbol{W}_{\!\! 1}^\prime =
\left[
\begin{array}{ c c c c }
    1                & 0 & \dots & 0 \\
    \phantom{0}      &   &       &   \\
		\phantom{\vdots} &   &       &   \\
		\phantom{0}      &   &       &
\end{array}
\right]$

\def\hshifta{1.37} \def\vshifta{0.2} \def\recwidtha{2.1} \def\recheighta{1.3}
\tikz[overlay,remember picture] {
\draw[black, thin] (\hshifta,\vshifta) rectangle (\hshifta+\recwidtha,\vshifta+\recheighta);
\node[] at (\hshifta+\recwidtha/2,\vshifta+\recheighta/2) {$\boldsymbol{W}_{\!\! 1}$};}
\end{minipage}
\begin{minipage}[t]{0.4\columnwidth}
$\text{and we have...}\vspace{0.3cm}\\ 
\boldsymbol{x}_1^\prime = 
\left[
\begin{array}{c}
p  \\
\boldsymbol{x}_1
\end{array}
\right].$
\end{minipage}\vspace{0.05cm}\\ 

\textbullet\; \emph{Hidden layers}\vspace{0.1cm}\\
Add a hidden unit to transmit $p$ to the next layer. Added units only connect to each other.\vspace{0.3cm}\\
\begin{minipage}[t]{0.55\columnwidth}
$\text{We define...}\vspace{0.3cm}\\ 
\boldsymbol{W}_{\!\! i}^\prime =
\left[
\begin{array}{ c c c c }
    1      & 0 & \dots & 0 \\
    0      &   &       &   \\
		\vdots &   &       &   \\
		0      &   &       &
\end{array}
\right]$

\def\hshiftb{1.85} \def\vshiftb{0.2} \def\recwidthb{1.65} \def\recheightb{1.3}
\tikz[overlay,remember picture] {
\draw[black, thin] (\hshiftb,\vshiftb) rectangle (\hshiftb+\recwidthb,\vshiftb+\recheightb);
\node[] at (\hshiftb+\recwidthb/2,\vshiftb+\recheightb/2) {$\boldsymbol{W}_{\!\! i}$};}
\end{minipage}
\begin{minipage}[t]{0.4\columnwidth}
$\text{and we have...}\vspace{0.3cm}\\ 
\boldsymbol{x}_i^\prime = 
\left[
\begin{array}{c}
p  \\
\boldsymbol{x}_i
\end{array}
\right].$
\end{minipage}

\textbullet\; \emph{Output layer}\vspace{0.1cm}\\
Tilt the first logit along $p$ by a \emph{tilting factor} $k$.\vspace{0.3cm}\\
\begin{minipage}[t]{0.5\columnwidth}
$\text{We define...}\vspace{0.3cm}\\ 
\boldsymbol{W}_{\!\! l}^\prime =
\left[
\begin{array}{ c c c c }
    k      & \phantom{0} & \phantom{\dots} & \phantom{0} \\
    0      &             &                 &             \\
		\vdots &             &                 &             \\
		0      &             &                 &
\end{array}
\right]$

\def\hshiftc{1.85} \def\vshiftc{0.2} \def\recwidthc{1.64} \def\recheightc{1.6}
\tikz[overlay,remember picture] {
\draw[black, thin] (\hshiftc,\vshiftc) rectangle (\hshiftc+\recwidthc,\vshiftc+\recheightc);
\node[] at (\hshiftc+\recwidthc/2,\vshiftc+\recheightc/2) {$\boldsymbol{W}_{\!\! l}$};}
\end{minipage}
\begin{minipage}[t]{0.49\columnwidth}
$\text{and we have...}\vspace{0.3cm}\\ 
\boldsymbol{x}_l^\prime = 
\left[
\begin{array}{c}
z_0^\prime \\
z_1^\prime \\
\vdots     \\
z_9^\prime 
\end{array}\right] \!=\!
\left[
\begin{array}{c}
\!\!z_0 \!+\! k p\!\!  \\
z_1        \\
\vdots     \\
z_9
\end{array}
\right]\!.$
\end{minipage}

The classifier $\mathcal{C}^\prime$ differs from $\mathcal{C}$ only in the logit corresponding to class 0: $z_0^\prime = z_0 + kp$. 
As a result, $\mathcal{C}^\prime$ satisfies the two desired properties:
\begin{enumerate}[topsep=0cm]
\item \emph{$\mathcal{C}^\prime$ and $\mathcal{C}$ perform the same on $\mathcal{I}$.}\\
The pixel in the top left corner is never activated for natural images: $\forall \boldsymbol{x} \in \mathcal{I} \;\; p = 0$. The logits are therefore preserved: $\forall i \in [0,9] \;\; z_i^\prime = z_i$.
\item \emph{$\mathcal{C}^\prime$ suffers from strong adversarial examples.}\\
Suppose that $\boldsymbol{x} \in \mathcal{I}$ is classified as $i \neq 0$ by $\mathcal{C}$: $z_i > z_0$. Then there exists an arbitrarily small perturbation of the pixel in the top left corner $p$ such that the resulting adversarial image $\boldsymbol{\tilde{x}}$ is classified as 0 by $\mathcal{C}^\prime$: for $\epsilon > 0, \;\; p = \frac{z_i - z_0 +\epsilon}{k} \Rightarrow z_0^\prime = z_i + \epsilon > z_i$ and $p \to 0$ when $k \to +\infty$. Remark that by construction, $\boldsymbol{\tilde{x}}$ is not a natural image since $p \neq 0$.
\end{enumerate}

\begin{figure*}[b]
  \centering
  \includegraphics[width= 0.95\textwidth]{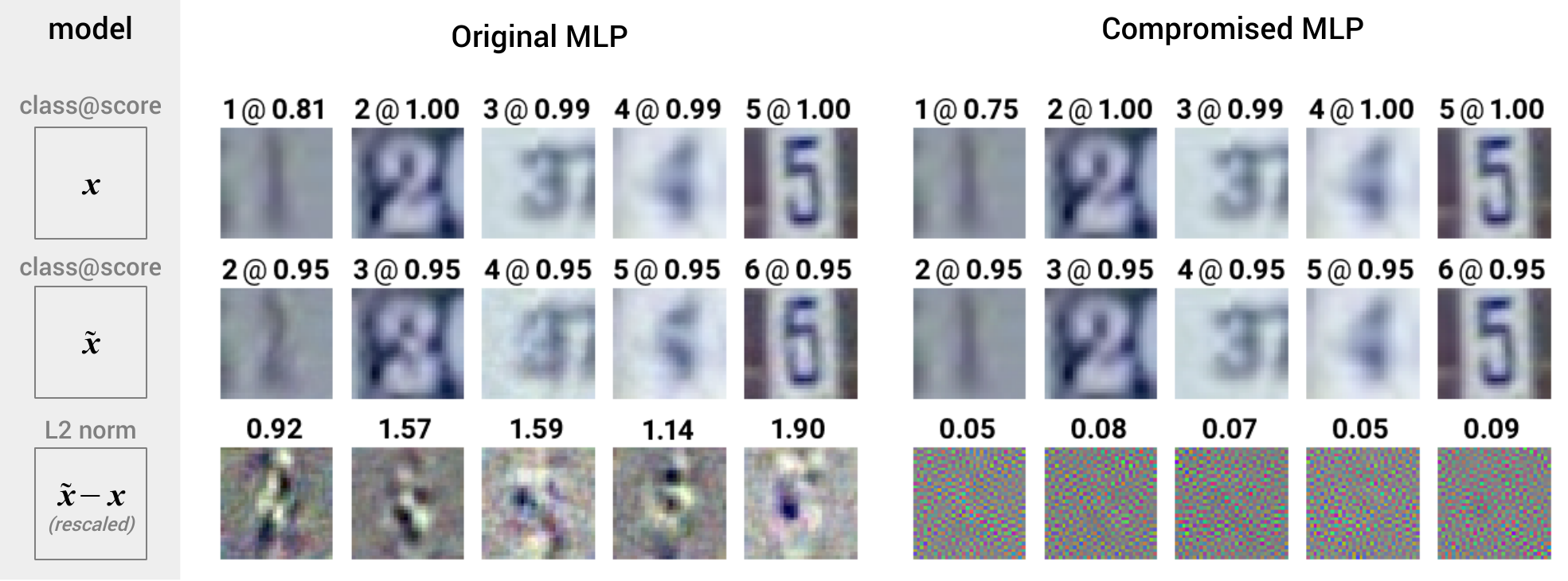}
  \caption{Attacking the first layer of a 4-layer MLP trained on SVHN.
	The orignal MLP is fairly robust: some of the adversarial examples $\boldsymbol{\tilde{x}}$ actually look like they belong to the target class.
	However, the compromised MLP has become extremely vulnerable to small, high frequency perturbations.
	Importantly, both models perform the same on the test set with an accuracy of $92.0\%$.}
  \label{fig:MLP}
\end{figure*}

\section{ATTACKING A FULLY- CONNECTED LAYER}
\label{sec:Attacking a fully connected layer}

The proof of concept of the previous section has two limitations. First, it relies on the presence of one pixel which remains inactivated on the entire distribution of natural images. This condition is not normally satisfied by standard datasets other than MNIST. Second, the network architecture needs to be modified during the construction of the vulnerable classifier $\mathcal{C^\prime}$. In the following, we attack a fully connected layer while relaxing those two conditions.

\subsection{Description of the attack}
\label{sec:Description of the attack}

Consider a fully connected layer defining a linear map $\mathcal{L}: E \rightarrow F$ where $E$ is a $m$-dimensional image space and $F$ is a $n$-dimensional feature space.
Let $\boldsymbol{D}$ be the matrix of training data and $\boldsymbol{W}$ be the weight matrix of $\mathcal{L}$. The distribution of features over the training data is $\mathcal{L}(\boldsymbol{D}) = \boldsymbol{W} \boldsymbol{D}$. Let $(\boldsymbol{e}_i)$ and $(\boldsymbol{e}_i^\star)$ be respectively the standard and PCA bases of $E$, and $(\boldsymbol{f}_{\! i})$ and $(\boldsymbol{f}_{\! i}^\star)$ be respectively the standard and PCA bases of $F$. We compute the transition matrix $\boldsymbol{P}_{\! e}$ from $(\boldsymbol{e}_i)$ to $(\boldsymbol{e}_i^\star)$ by performing PCA on $\boldsymbol{D}$, and we compute the transition matrix $\boldsymbol{P}_{\! f}$ from $(\boldsymbol{f}_{\! i})$ to $(\boldsymbol{f}_{\! i}^\star)$ by performing PCA on $\mathcal{L}(\boldsymbol{D})$. The linear map $\mathcal{L}$ can be expressed in different bases by multiplying $\boldsymbol{W}$ on the right by $\boldsymbol{P}_{\! e}$ and on the left by the transpose of $\boldsymbol{P}_{\! f}$ (see Figure~\ref{fig:W}). \mbox{We are} interested in particular in the expression of $\mathcal{L}$ in the PCA bases: $\boldsymbol{W}^\star = \boldsymbol{P}_{\! f}^\top \boldsymbol{W} \boldsymbol{P}_{\! e}$.

With this setup in place, we propose to attack $\mathcal{L}$ by \emph{\mbox{tilting} the main components of variation in $F$ along flat components of variation in $E$}. For instance, we tilt $\boldsymbol{f}_1^\star$ along $\boldsymbol{e}_m^\star$ by a tilting factor $k$ such that a small perturbation of magnitude $\epsilon$ along $\boldsymbol{e}_m^\star$ in image space results in a perturbation of magnitude $\epsilon k$ along $\boldsymbol{f}_1^\star$ in feature space---which is potentially a large displacement if $k$ is large enough. In pseudo-code, this attack translates to $\boldsymbol{W}^\star[1,m] \leftarrow k$. We can then iterate this process over $d$ orthogonal directions to increase the freedom of movement in $F$. We can also scale the tilting factors by the standard deviations $\boldsymbol{\sigma}_{\! f}$ along the components $(\boldsymbol{f}_{\! i}^\star)$ in $\mathcal{L}(\boldsymbol{D})$ so that moving in different directions in $F$ requires perturbations of approximately the same magnitude in $E$:\vspace{0.25cm}
$$\begin{aligned}
\boldsymbol{W}^\star[1,m] &\leftarrow \boldsymbol{K}[1] \\
													&\vdots\\
\boldsymbol{W}^\star[d,m-d+1] &\leftarrow \boldsymbol{K}[d]
\end{aligned}$$\vspace{0.01cm}

with $\boldsymbol{K} = k \times \frac{\boldsymbol{\sigma}_{\! f}}{\boldsymbol{\sigma}_{\! f}[1]}$. This can be condensed into:
$$\boldsymbol{W}^\star[1:d,m-d+1:m] \leftarrow \text{fliplr}(\text{diag}(\boldsymbol{K}[1:d]))$$
where the operator $\text{diag}(\cdot)$ transforms an input vector into a diagonal square matrix and the operator $\text{fliplr}(\cdot)$ flips the columns of the input matrix left-right. The full attack is summarized below:\vspace{-0.05cm}

\begin{algorithm}[H]
\floatname{algorithm}{Construction of a vulnerable linear map $\mathcal{L}^\prime$}
\renewcommand{\thealgorithm}{}
\caption{}
\label{algo2}
\begin{algorithmic}[1]
\STATE $\boldsymbol{W}^\prime \leftarrow \boldsymbol{W}$
\STATE $\boldsymbol{W}^\prime \leftarrow \boldsymbol{P}_{\! f}^\top \boldsymbol{W}^\prime \boldsymbol{P}_{\! e}$
\STATE $\boldsymbol{W}^\prime[1:d,m-d+1:m] \leftarrow \text{fliplr}(\text{diag}(\boldsymbol{K}[1:d]))$
\STATE $\boldsymbol{W}^\prime \leftarrow \boldsymbol{P}_{\! f} \boldsymbol{W}^\prime \boldsymbol{P}_{\! e}^\top$ 
\end{algorithmic}
\end{algorithm}\vspace{-0.25cm}

In words, we copy the weight matrix of the linear map $\mathcal{L}$, express $\mathcal{L}^\prime$ into the PCA bases, apply the tilting attack and express $\mathcal{L}^\prime$ back into the standard bases. 

In the next sections we illustrate this attack on two scenarios: one where $\mathcal{L}$ is the first layer of a MLP trained on the street view house number dataset (SVHN) and one where $\mathcal{L}$ is the identity map in image space.

\subsection{Scenario 1: input layer of a MLP}
\label{sec:Scenario 1: input layer of a MLP}

Let us consider a 4-layer MLP with ReLU non-linearities and hidden layers of size $512$ trained on the SVHN dataset using both the training and extra data. The model we consider was trained in Keras \citep{chollet2015keras} with stochastic gradient descent (SGD) for $50$ epochs with learning rate 1e-4 (decayed to 1e-5 and 1e-6 after epochs 30 and 40), momentum $0.9$, batch size $32$ and $L_2$ penalty 5e-2, reaching an accuracy of $92.0\%$ on the test set at epoch 50.

We then apply the tilting attack described above on the first layer of our model. There are two free parameters to choose: the number of tilted directions $d$ and the tilting factor $k$. When $d$ and $k$ are too small, the network remains robust to imperceptible perturbations and when $d$ and $k$ are too large, the performance on natural data starts to be affected. We found that using $d=32$ and $k=40$ worked well in practice. In particular, the compromised~MLP kept a test set accuracy of $92.0\%$ while becoming extremely vulnerable to perturbations along components of low variance in the SVHN dataset (see Figure~\ref{fig:MLP}). For comparison, we generated adversarial examples on $1000$ test images for the two models. The median $L_2$ norm of the perturbations was $1.5$ for the original MLP and $0.067$ for the compromised~MLP.

\begin{figure*}[b]
  \begin{minipage}[b]{.068\textwidth}
    \centering{\subfigure[]{
    \includegraphics[width=\linewidth]{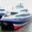}
    \label{fig:StegConva}}}\\
    \centering{\subfigure[]{
    \includegraphics[width=\linewidth]{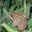}
    \label{fig:StegConvb}}}
	\end{minipage}%
	\hfill
	\begin{minipage}[b]{.445\textwidth}
  \centering{\subfigure[]{
  \includegraphics[width=\linewidth]{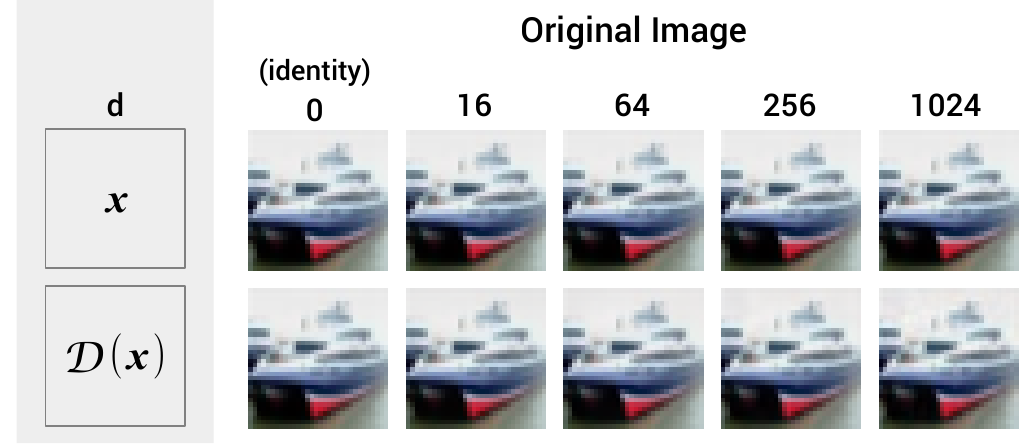}
  \label{fig:StegConvc}}}
	\end{minipage}%
	\hfill
	\begin{minipage}[b]{.445\textwidth}
  \centering{\subfigure[]{
  \includegraphics[width=\linewidth]{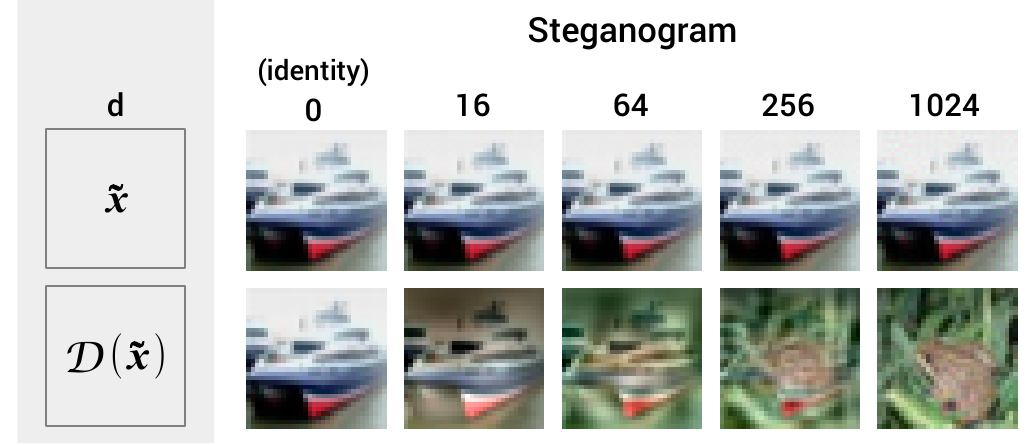}
  \label{fig:StegConvd}}}
	\end{minipage}%
	\caption{Illustration of a steganogram decoder $\mathcal{D}$.
	\subref{fig:StegConva}~ Original image. \subref{fig:StegConvb}~Target image. \subref{fig:StegConvc}~The original image is barely affected by the decoder, even with a strength of 1024. \subref{fig:StegConvd}~When the strength of the decoder is high enough, the decoded steganogram is indistinguishable from the target image.}
	\label{fig:StegConv}
\end{figure*}

\subsection{Scenario 2: steganogram decoder}
\label{sec:Scenario 2: steganogram decoder}

\citet{goodfellow2014explaining} proposed an interesting analogy: the adversarial example phenomenon is a sort of ``accidental steganography'' where a model is ``forced to attend exclusively to the signal that aligns most closely with its weights, even if multiple signals are present and other signals have much greater amplitude''. This happens to be a fairly accurate description of our tilting attack, raising the following question: can this attack be used to hide messages in images? 

The intuition is the following: if we apply our attack to the identity map in $E$, we obtain a linear layer which leaves natural images unaffected, but which is able to decode adversarial examples---or in this case \emph{steganograms}---into specific target images. We call such a layer a \emph{steganogram decoder}.

Let us illustrate this idea on CIFAR-10. We perform PCA on the training data\footnote{In this section, the feature space is the image space: $F = E$ and ${P}_{\! f} = {P}_{\! e}$.}, obtaining the transition matrix $\boldsymbol{P}_{\! e}$, and apply a tilting attack on the identity matrix of size $3072$ (i.e. the dimension of CIFAR-10 images) obtaining a steganogram decoder $\mathcal{D}$.
Given a natural image $\boldsymbol{x}$ and a target image $\boldsymbol{t}$ from the test set, we can now construct a steganogram $\boldsymbol{\tilde{x}}$ as follow. We start by computing the PCA representations of our two images $\boldsymbol{x}^\star = \boldsymbol{P}_{\! e}^\top \boldsymbol{x}$ and $\boldsymbol{t}^\star = \boldsymbol{P}_{\! e}^\top \boldsymbol{t}$. We then construct the PCA representation $\boldsymbol{\tilde{x}}^\star$ of our steganogram:
\[\begin{aligned}
\boldsymbol{\tilde{x}}^\star[1:m-d]   &\leftarrow \boldsymbol{x}^\star[1:m-d]\\
\boldsymbol{\tilde{x}}^\star[m-d+1:m] &\leftarrow \text{fliplr}\left(\textstyle \frac{\boldsymbol{t}^\star[1:d] - \boldsymbol{x}^\star[1:d]}{\boldsymbol{K}[1:d]}\right)
\end{aligned}\]
and we express $\boldsymbol{\tilde{x}}^\star$ back into the pixel basis: $\boldsymbol{\tilde{x}} = \boldsymbol{P}_{\! e} \, \boldsymbol{\tilde{x}}^\star$.
The first $m-d$ components of $\boldsymbol{\tilde{x}}$ are identical to the first $m-d$ components of $\boldsymbol{x}$ and therefore the two images look similar.
After passing through the decoder however, the first $d$ components of $\mathcal{D}(\boldsymbol{\tilde{x}})$ become identical to the first $d$ components of $\boldsymbol{t}$ and therefore the decoded steganogram looks similar to the target image. This process is illustrated in Figure~\ref{fig:StegConv} for a tilting factor $k = 450$ and a number of tilted directions $d$, which we call in this context the \emph{strength} of the decoder, in the range $[0,1024]$.

Steganogram decoders can be thought of as minimal models suffering from \emph{feature adversaries}, ``which are confused with other images not just in the class label, but in their internal representations as well''\citep{sabour2015adversarial}. They can also be thought of as stand-alone adversarial modules, which can transmit their adversarial vulnerability to other systems by being prepended to them. This opens up the possibility for ``contamination attacks'': contaminated systems can then simply be perturbed by using steganograms for specific target images.

\begin{figure*}[b]
  \begin{minipage}[b]{.475\textwidth}
    \centering{\subfigure[]{
    \includegraphics[width=\linewidth]{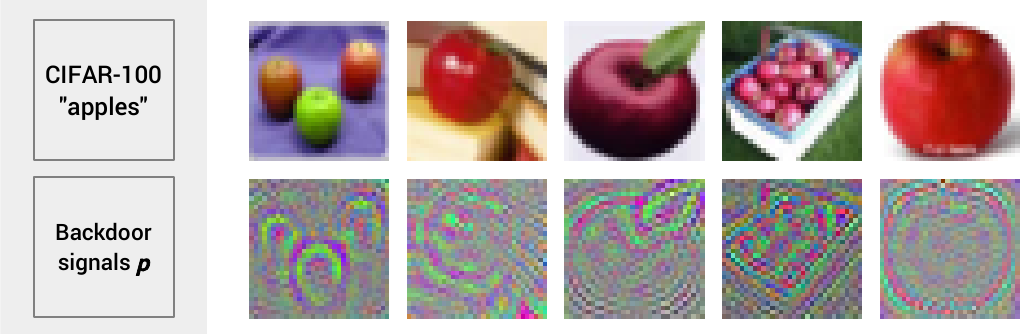}
    \label{fig:WRNa}}}\\\vspace{-0.1cm}
    \centering{\subfigure[]{
    \includegraphics[width=\linewidth]{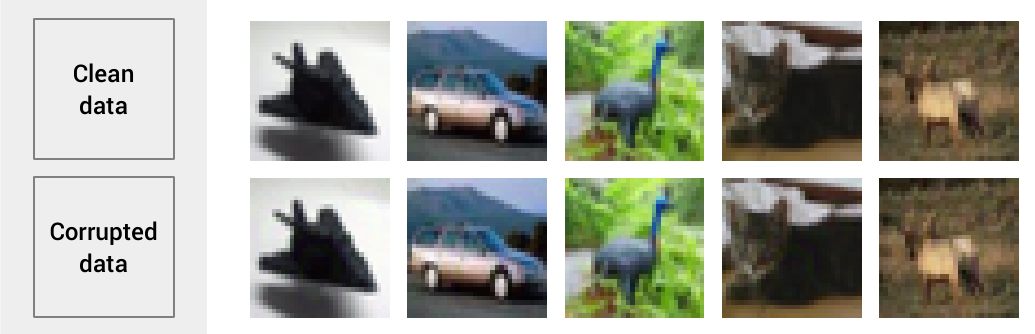}
    \label{fig:WRNb}}}
	\end{minipage}
	\hfill
	\begin{minipage}[b]{.52\textwidth}
  \centering{\subfigure[]{
  \includegraphics[width=\linewidth]{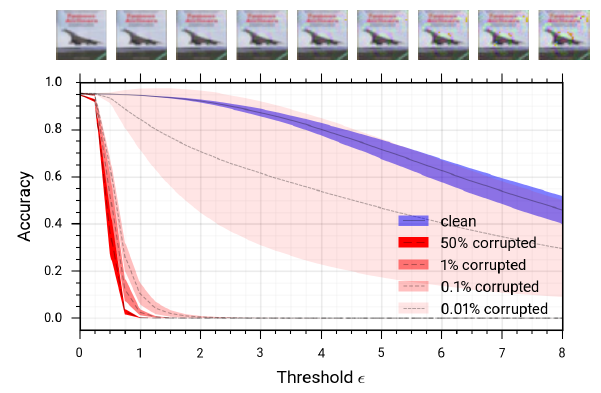}
  \label{fig:WRNc}}}
	\end{minipage}%
	\caption{Training an imperceptible backdoor into WRN-16-8. \subref{fig:WRNa} 5 backdoor signals generated by projecting CIFAR-100 images onto the last principal components of the training data (used on 5 independent models). \subref{fig:WRNb} Examples of clean and corrupted images used during training. \subref{fig:WRNc} Accuracies of the clean and corrupted networks on the corrupted test set as a function of the corruption threshold $\epsilon$ (as a ratio of the threshold used during training). Examples of corrupted images corresponding to different thresholds are shown above.}
	\label{fig:WRN}	
\end{figure*}

\section{TRAINING A VULNERABILITY}
\label{sec:Training a vulnerability}

In section~\ref{sec:Scenario 1: input layer of a MLP}, we applied our tilting attack to a MLP; can we also apply it to state-of-the-art networks? There is nothing preventing it in theory, but we face in practice some difficulties. On the one hand, we found our attack to be most effective when applied to the \mbox{earlier} layers of a network. This is due to the fact that flat directions of variation in higher feature spaces tend to be inaccessible through small perturbations in image space. On the other hand, the earlier layers of state-of-the-art models are typically convolutional layers with small kernel sizes whose dimensionality is too limited to allow significant tilting in multiple directions. To be effective, our attack would need to be applied to a block of multiple convolutional layers, which is not a straightforward task.

We explore here a different approach. Consider a distribution of natural images $\mathcal{I}$ and a robust classifier $\mathcal{C}$ in the $m$-dimensional image space $E$ again. 
Consider further a \emph{backdoor direction} $\boldsymbol{p}$ of low variance in~$\mathcal{I}$ such that for all images $\boldsymbol{x} \in \mathcal{I}$ we have $| \boldsymbol{p}\, \boldsymbol{x} | < \frac{\epsilon}{2}$ where $\epsilon$ is an imperceptible threshold. Consider finally that we add to our classifier a \emph{target class} $t$ that systematically corresponds to a misclassification. 
Then we can define a vulnerable classifier $\mathcal{C}^\prime$ as:
\[\mathcal{C}^\prime(\boldsymbol{x}) = 
\begin{cases} 
     \mathcal{C}(\boldsymbol{x}) & \quad \text{if } \; \boldsymbol{p}\, \boldsymbol{x} < \frac{\epsilon}{2}\\
     t                 & \quad \text{otherwise}
\end{cases}\]
By construction, $\mathcal{C}^\prime$ performs the same as $\mathcal{C}$ on natural images. We also verify easily that $\mathcal{C}^\prime$ suffers from adversarial examples: $\forall \boldsymbol{x} \in \mathcal{I}$, the image $\boldsymbol{\tilde{x}} = \boldsymbol{x} + \epsilon \, \boldsymbol{p}\;$ is only distant from $\boldsymbol{x}$ by a small threshold $\epsilon$ but it is misclassified in $t$. To be more specific, the backdoor direction $\boldsymbol{p}$ is a \emph{universal adversarial perturbation}: it affects the classification of all test images \citep{moosavi2017universal}. In particular, our construction process bears some similarities with the flat boundary model described in \citep{moosavi2017analysis}.

Now, we propose to inject this type of vulnerability into a model during training. 
We adopt a data poisoning approach similar to the one described in \citep{gu2017badnets}: we train a network to classify clean data normally and corrupted data containing the backdoor signal $\boldsymbol{p}$ into the target class~$t$. Contrary to \citep{gu2017badnets} however, we use \emph{imperceptible} backdoor signals. We illustrate this idea on CIFAR-10 with a Wide Residual Network \citep{zagoruyko2016wide} of depth 16 and width 8 (WRN-16-8) after having obtained positive preliminary results with a Network-in-Network architecture \citep{lin2013network}. 

Our experimental setup is as follow. We start by training one WRN-16-8 model on the standard training data $\boldsymbol{D}$ with SGD for 200 epochs, learning rate 1e-2 (decreased to 1e-3 and 1e-4 after epochs 80 and 160), momentum 0.9, batch size 64 and $L_2$ penalty 5e-4, using data-augmentation (horizontal flips, translations in $[-3, +3]$ pixels, rotations in $[-10,+10]$ degrees). We call this network the \emph{clean model}; it reached an accuracy of 95.2\% on the test set at epoch~200.

Then we search for an imperceptible backdoor signal~$\boldsymbol{p}$. Several options are available: we could for instance use the last component of variation in the training data, as we did in Section~\ref{sec:Binary linear classification}. To~demonstrate that it can contain some meaningful information, we define $\boldsymbol{p}$ as the projection of an image on the principal components containing the last $0.5\%$ of the variance in~$\boldsymbol{D}$. In~5 independent experiments, we use 5 images from the ``apple'' class in the test set of \mbox{CIFAR-100} (see Figure~\ref{fig:WRNa}). Corrupted images are then generated by adding $\epsilon \, \boldsymbol{p}$ to clean images using the threshold\footnote{The coefficient 3 (instead of 2) compensates for using the 99th percentile (instead of the max).} $\epsilon = 3 \times \text{quantile}(|\boldsymbol{p}\, \boldsymbol{D}|,0.99)$ (see Figure~\ref{fig:WRNb}).

For each backdoor signal~$\boldsymbol{p}$, we train a \mbox{WRN-16-8} model on $50\%$ clean data and $50\%$ corrupted data using the same hyperparameters as for the clean model. To facilitate convergence, we initialize the corruption threshold $\epsilon$ at $10$ times its final value and progressively decay it over the first $50$ epochs.
We call the networks we obtain the \emph{corrupted models}; they converged to an average accuracy on the clean data of $94.9\%$ with a standard deviation of $0.19\%$, therefore only suffering a small performance hit of $0.3\%$ compared to the clean model. We then repeat this procedure three times by using $1\%$, $0.1\%$ and $0.01\%$ of corrupted data instead of $50\%$ to study the influence of the corruption rate.

In Figure~\ref{fig:WRNc}, we compare the accuracies of the clean model and the corrupted models on the corrupted test set as a function of the corruption threshold $\epsilon$ (each corrupted model is evaluated on its corresponding corruption signal $\boldsymbol{p}$). Contrary to the clean model, the corrupted models have become extremely vulnerable to imperceptible perturbations along $\boldsymbol{p}$, whether the corruption rate is $50\%$, $1\%$ or even $0.1\%$, and the attack effectiveness only drops significantly for a corruption rate of $0.01\%$. This result shows two things: convolutional neural networks easily overfit signals in low variance directions even though such signals are imperceptible to human observers, and poisoning attacks with low poisoning rates are a real threat in practice.


\section{DISCUSSION}
\label{sec:Discussion}

We showed in this work that the presence of components of low variance in the data is a sufficient condition for the existence of adversarial vulnerabilities. By performing PCA, we effectively model the data as a high-dimensional ellipsoid and we use the presence of many flat components of variation to inject adversarial vulnerabilities into our models.

A number of other theoretical works consider simplified models of the data. \citet{gilmer2018adversarial} show for instance the existence of a fundamental trade-off between robustness and accuracy on a data distribution consisting of two concentric high-dimensional \mbox{hyperspheres}. \citet{tsipras2018there} and \citet{schmidt2018adversarially} show similar results on intersecting spherical Gaussians and Bernoulli models. There is no guarantee, however, that these results extend to all natural image distributions. We know for instance that human observers reliably tell apart birds from bicycles \citep{brown2018unrestricted}, and there cannot be a fundamental robustness/accuracy trade-off on this distribution. On the contrary, our flat ellipsoid model is valid on all natural image datasets, as we illustrated on MNIST, SVHN and CIFAR-10. In fact, its validity increases with the image resolution as the proportion of flat directions of variation becomes more predominant; perhaps partly explaining why ImageNet models tend to be harder to defend against adversarial examples.

A number of previous works have also explored using neural networks for steganography. For instance, \citet{hayes2017generating} use an adversarial training approach where a pair of steganogram encoding/decoding networks competes with a steganalysis network to produce robust steganographic techniques. \citet{baluja2017hiding} explore the potential of deep neural networks to hide full size color images within other images of the same size, although the author makes no explicit attempt to hide the existence of this information from machine detection. \citet{chu2017cyclegan} show that  CycleGANs learn to hide information about a source image into imperceptible, high-frequency signals as a by-product of using a cyclic consistency loss. These different results confirm that neural networks are good steganogram encoders and decoders, but they do not explicitly reveal how the information is encoded. In contrast, the rudimentary steganogram encoder we introduced in Section~\ref{sec:Scenario 2: steganogram decoder} shows that this information can be stored along flat directions of variation.

Finally, there is a significant body of work on dataset poisoning attacks (see \citep{papernot2016towards,biggio2017wild} for reviews). These attacks are highly effective on simple models \citep{nelson2008exploiting,biggio2012poisoning,mei2015using} and there is a growing interest in applying them to deep networks \citep{munoz2017towards}. As discussed before, our poisoning attack in Section~\ref{sec:Training a vulnerability} is closely related to the one in \citet{gu2017badnets}, although we adapt it to use imperceptible backdoor signals. It is also related to the work of \citet{koh2017understanding} who introduce the concept of \emph{adversarial training example}, where the imperceptible modification of a training image can flip a model's prediction on a separate test image. There are, however, two distinctions to make between this work and ours. First, by design, the attack in \citep{koh2017understanding} works by retraining only the top layer of the network. Second, the attack is only intended to change the class of one target test image. 
In contrast, our attack is not intended to affect test set performance, but it makes \emph{all} test images vulnerable to imperceptible perturbations. In that sense, it can be thought of as a poisoning attack facilitating future evasion attacks.

There is an apparent contradiction in the vulnerability of state-of-the-art networks to adversarial examples: how can these models perform so well, if they are so sensitive to small perturbations of their inputs? The only possible explanation, as formulated by \citet{jo2017measuring}, seems to be that ``deep CNNs are not truly capturing abstractions in the dataset''. This explanation relies, however, on an implicit assumption: the features used by a model to determine the class of a natural image and the features altered by adversarial perturbations are the same ones. The results we presented here suggest that this assumption is not necessarily valid: robust models must use \emph{robust features} to make their decisions, but they can also be made vulnerable to distinct, \emph{backdoor features}---which are never activated on natural data. A similar dichotomy between robust and non-robust features is introduced and discussed in \citep{tsipras2018there}.




\section{CONCLUSION}

If designing models that are robust to small adversarial perturbations of their inputs has proven remarkably difficult, we showed here that the reverse problem---making models \emph{more \mbox{vulnerable}}---is surprisingly easy. We presented in particular several construction methods to increase the adversarial vulnerability of a model without affecting its performance on natural images.

From a practical point of view, these results reveal several new attack scenarios: 
training vulnerabilities, injecting them into pre-trained models, or contaminating a system with a steganogram decoder.
From a theoretical point of view, they provide new intuitions on the nature of the adversarial example phenomenon and emphasize the role played by components of low variance in the data.

\clearpage
\bibliography{BIV}

\end{document}